\documentclass[pmlr,twocolumn,10pt]{jmlr} 

\usepackage{booktabs}
\usepackage{siunitx}
\usepackage{smile}

\usepackage[switch]{lineno}

\usepackage{pmboxdraw}
\usepackage{xcolor, colortbl}  

\newcommand{\ours}{\texttt{HTP-Star}\xspace}

\theorembodyfont{\upshape}
\theoremheaderfont{\scshape}
\theorempostheader{:}
\theoremsep{\newline}

\jmlrvolume{LEAVE UNSET}
\jmlryear{2024}
\jmlrsubmitted{LEAVE UNSET}
\jmlrpublished{LEAVE UNSET}
\jmlrworkshop{Conference on Health, Inference, and Learning (CHIL) 2024} 

\title[HTP-Star]{From Basic to Extra Features: Hypergraph Transformer Pretrain-\\then-Finetuning for Balanced Clinical Predictions on EHR}

\author{%
\Name{Ran Xu}\Email{ran.xu@emory.edu}\\
\addr Emory University, United States
\AND
\Name{Yiwen Lu} \Email{yiwenlu@sas.upenn.edu}\\
\addr University of Pennsylvania, United States
\AND
\Name{Chang Liu} \Email{chang.liu2@emory.edu}\\
\addr Emory University, United States
\AND
\Name{Yong Chen} \Email{ychen123@pennmedicine.upenn.edu}\\
\addr University of Pennsylvania, United States
\AND
\Name{Yan Sun} \Email{yan.v.sun@emory.edu}\\
\addr Emory University, United States
\AND
\Name{Xiao Hu} \Email{xiao.hu@emory.edu}\\
\addr Emory University, United States
\AND
\Name{Joyce C Ho} \Email{joyce.c.ho@emory.edu}\\
\addr Emory University, United States
\AND
\Name{Carl Yang} \Email{j.carlyang@emory.edu}\\
\addr Emory University, United States
}

\begin{document}

\maketitle

\begin{abstract}
Electronic Health Records (EHRs) contain rich patient information and are crucial for clinical research and practice. 
In recent years, deep learning models have been applied to EHRs, but they often rely on massive features, which may not be readily available for all patients.
We propose \ours{}\footnote{Short for \textbf{H}ypergraph \textbf{T}ransformer \textbf{P}retrain-then-Finetuning with \textbf{S}moo\textbf{t}hness-induced regularization \textbf{a}nd \textbf{R}eweighting.}, which leverages hypergraph structures with a pretrain-then-finetune framework for modeling EHR data, enabling seamless integration of additional features. 
Additionally, we design two techniques, namely
(1) \emph{Smoothness-inducing Regularization} and 
(2) \emph{Group-balanced Reweighting},
to enhance the model's robustness during finetuning. 
Through experiments conducted on two real EHR datasets, we demonstrate that \ours{} consistently outperforms various baselines while striking a balance between patients with basic and extra features.
\end{abstract}

\paragraph*{Data and Code Availability}
We evaluate our framework on two publicly available datasets UK Biobank~\citep{sudlow2015uk} and MIMIC-III~\citep{johnson2016mimic}.
The research was conducted using data from the UK Biobank Resource under an application number (omitted for anonymization). 
The UK Biobank makes the data available to all bona fide researchers for all types of health-related research that is in the public interest, without preferential or exclusive access for any persons. All researchers are subject to the same application process and approval criteria as specified by UK Biobank.
MIMIC-III is publicly available from the PhysioNet repository.

\paragraph*{Institutional Review Board (IRB)}
UK Biobank has approval from the North West Multi-centre Research Ethics Committee (MREC) as a Research Tissue Bank (RTB) approval. MIMIC-III does not need IRB approval.

\section{Introduction}
\label{sec:intro}

Electronic Health Record (EHR) is a digital representation of a patient's medical history that contains a wealth of patient information, including diagnoses, medications, lab results, and so on~\citep{cowie2017electronic}.
In clinical research and practice, healthcare professionals actively use EHRs for patient health monitoring~\citep{gandrup2020remote,shi2024ehragent}, risk predictions~\citep{pmlr-v209-la-cava23a} and clinical trial matching~\citep{rogers2021comparison}, thereby harnessing the capabilities of digital repositories to augment patient care and inform clinical decision-making~\citep{tang2022leveraging}.

\begin{figure}[t]
\floatconts
  {fig:hypergraph_intro}
  {\caption{An illustration of basic and extra features.}\vspace{-1.5ex}}
  {\includegraphics[width=0.95\linewidth]{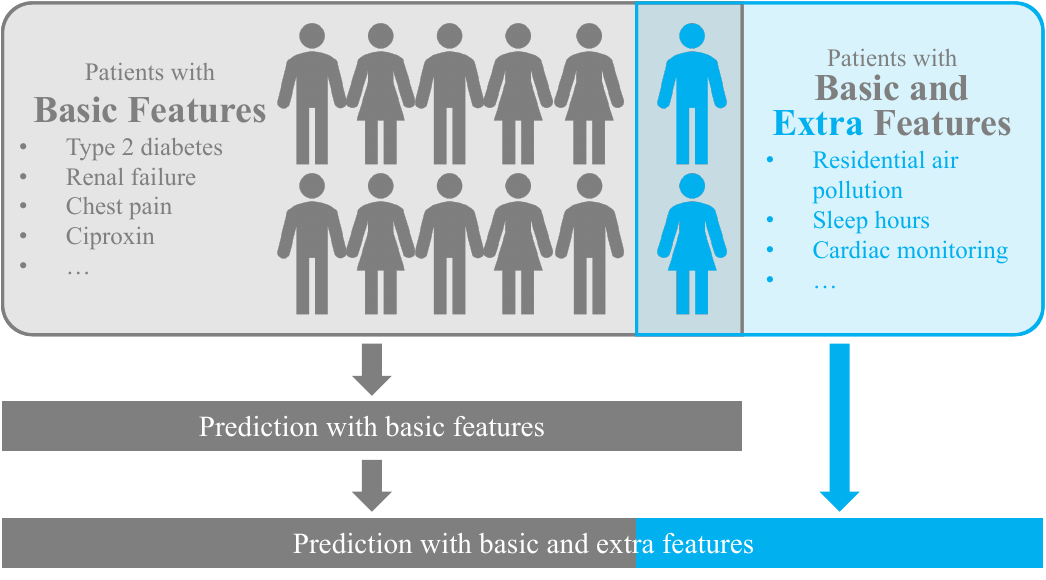}\vspace{-3ex}}
  \vspace{-3ex}
\end{figure}

In recent years, various deep learning architectures~\citep{choi2016retain,pang2021cehr} have gained extensive popularity in predictive tasks on EHR. 
Typically, these models are trained on a collection of basic features shared across various medical institutions, such as diseases and medications.
However, in real-world scenarios, additional features are often collected, which are limited to specific hospitals due to privacy or budget constraints~\citep{taksler2021opportunities,hong2021clinical}. 
Figure~\ref{fig:hypergraph_intro} shows an illustration of extra features collected by some local medical institutes.
Due to the often small sample sizes for patients with these extra features, they are not effectively utilized to enhance the modeling of patients, 
regardless of whether they possess these extra features or not. 
Hence, our primary objective in this research is to address the challenge: \emph{How can we harness the extra data gathered from local medical institutes on specific patients to enhance clinical prediction tasks within the population?}

This naturally resembles a transfer learning setting~\citep{weiss2016survey} with a pretrain-then-finetune pipeline, which requires the model to effectively transfer the knowledge from a more extensive population to individuals with extra sets of features. 
However, developing clinical predictive models that concurrently incorporate basic and extra features is nontrivial. 
Directly adapting traditional machine learning (ML) methods such as linear regression or decision trees is problematic since the model trained from patients with basic features cannot directly handle patients with extra features due to different feature dimensions. 
Although there exist several deep learning architectures (e.g. Transformers~\citep{lee2019set,choi2020learning}) that can model flexible numbers of patient features, 
they do not explicitly model the interactions between features, so they cannot fully leverage extra features to improve the modeling of basic features, and vice versa.

Given the challenges outlined above, we emphasize the importance of designing a suitable data structure capable of accommodating additional features from local institutes for patients. 
Inspired by the recent progress in hypergraph learning for clinical predictions on EHRs with strong representative power~\citep{pmlr-v193-xu22a,cai2022hypergraph,wu2023megacare}, we propose \ours{} to utilize \emph{hypergraph} structure to characterize the EHR data. 
Building upon this structure, patient visits are conceptualized as hyperedges, with each visit-related feature represented as a node, allowing each hyperedge to be connected to a flexible number of nodes. 
This approach not only effectively characterizes the relationships between hospital visits and medical codes from a higher-order view, but also enables the seamless integration of new features into the current dataset by simply adding nodes to the existing hyperedges without extensive modifications to the overall graph structure. 
Additionally, to facilitate information propagation and mutual enhancement between newly incorporated and existing basic features, we employ \emph{hypergraph transformers}~\citep{pmlr-v193-xu22a,cai2022hypergraph}, which incorporates multi-head self-attention and jointly learns the embeddings for hospital visits and all patient features.

After capturing the relationship between visits and features via hypergraph transformer, it is also crucial to design \emph{effective and balanced training techniques} to enable models to generalize well on both basic and extra features. This is essential as samples with basic and extra features might sometimes have conflicts in their optimization directions.
Existing transfer learning models leverage self-supervised learning~\citep{shang2019pre,bo2022pretraining,pmlr-v174-park22a,mcdermott2021comprehensive} to improve the model's generalization ability with a pretrain-then-finetune pipeline, but they often directly fine-tune on target tasks without effective regularization, which is easy to cause catastrophic forgetting~\citep{ramasesh2021effect}.
There are also generic transfer learning methods~\citep{han2021adaptive,liu2021subtype,jiang2023forkmerge}, but they often have strong assumptions on data distributions, and thus may not adapt to the clinical setting well. 
Motivated by this, we develop two strategies to enhance the model's generalization ability across patients with varying data volumes: 
(1) To \emph{mitigate the risk of aggressive model updates}, we maintain a slowly updated predictive model, which takes the form of a momentum-based moving average of the originally fine-tuned model. We add a regularization term to encourage consistent predictions between the original and the slowly updated predictive model to prevent the predictive model from forgetting previous information learned from pretraining. 
(2) To \emph{resolve conflicts in optimization directions between basic and extra data features}, we introduce a gradient balancing method that adjusts the combination of gradients from patients with different data types.
With these two dedicated techniques, {\ours} learns a robust hypergraph model for EHR predictive tasks to accommodate both basic and extra features simultaneously.





We conduct experiments on two datasets, UKBiobank~\citep{sudlow2015uk} and MIMIC-III~\citep{johnson2016mimic}, to evaluate \ours{} and potential baselines.
The results demonstrate that \ours{} outperforms various standard ML methods as well as existing finetuning techniques, achieving a balance between patients with basic and extra features. Our contribution can be summarized as follows:
\begin{itemize}
    \item We study the problem of clinical predictions with basic and extra features and identify the challenges (Sec.~\ref{sec:prelim_limitation}), which have not been widely explored in prior works.
    \item We design \ours{}, a hypergraph pretrain-then-finetuning framework to enhance the model's robustness over two patient subgroups. We further propose two additional techniques to improve the model's generalization ability during fine-tuning steps.
    \item We conduct comprehensive experiments on two publicly available datasets (UK Biobank and MIMIC-III) to verify the efficacy of \ours{}.
\end{itemize}


\section{Related Works}
\label{sec:related}

\paragraph{Deep Predictive Model for EHRs}
In recent years, there have been numerous studies focusing on developing deep healthcare predictive models with various medical concepts. 
Earlier works attempt to leverage recurrent neural networks (RNN)~\citep{choi2016retain,lipton2016modeling} as well as Transformers~\citep{li2020behrt,pang2021cehr} to model the chronological relationships among different medical units. 
Graph-based models have also been proposed for EHR modeling, including graph convolution networks~\citep{zhu2021variationally,lu2022context}, 
graph transformers~\citep{choi2020learning,zhu2021variationally,jiang2023graphcare}, 
and hypergraph neural networks~\citep{cai2022hypergraph,pmlr-v193-xu22a}. 
These approaches involve constructing a co-occurrence graph based on EHR data and then using graph neural networks (GNNs) to learn the relations among medical codes within each for clinical outcome prediction~\citep{johnson2023graph}. 
Despite the impressive performance exhibited by deep learning-based models, these models typically demand massive labeled data and substantial feature richness, making them challenging to deploy in real-life resource-constrained healthcare environments~\citep{erion2022coai}.
In this study, we harness graph-based deep learning models coupled with enhanced training methodologies to address the challenge of data scarcity in EHRs. 
It is also worth noting that, unlike existing graph-based approaches which concentrate on enhancing performance for patients with only basic or extra features, our approach targets enhancing the generalization ability for patients with both \emph{basic} and \emph{extra} feature profiles.

\paragraph{Training Techniques for Better Generalization} 
Our work is also related to several studies for improving the model's generalization with basic data. 
\emph{Self-supervised learning} techniques has been widely adopted for CV and NLP tasks~\citep{devlin2019bert,chen2020simple}, 
and has also been adopted for EHRs with improved generalization~\citep{shang2019pre,mcdermott2021comprehensive,bo2022pretraining,pmlr-v174-park22a}.  
\emph{Transfer learning} techniques~\citep{weiss2016survey} aims to transfer knowledge across the target and source model, 
and recent works have proposed to harness attention networks~\citep{xiao2020cheer}, 
generative models~\citep{desautels2017using,estiri2021generative}, 
reweighting techniques~\citep{li2023multi,han2021adaptive,yu2021fine}, 
or adversarial networks~\citep{dai2022graph,liu2021subtype} to facilitate knowledge adaptation to the target domains.  
Our method is related as we first leverage self-supervised learning techniques to warm up the model training, and then design strategies to better transfer the knowledge to patients with both basic and extra features.


\section{Preliminary Studies}
\label{sec:prelim}
Before describing the details of our proposed model, we first give a brief overview of the problem setup, as well as the potential challenges under this scenario.

\subsection{Problem Setup}
In this study, we focus on predictive tasks on EHR which comprises patient visits with different medical codes. Formally, EHR visits are defined as:

\begin{definition}[EHR Visit]\label{def:EHR}
The EHR system generally includes a large amount of hospital visits $\cH$ for corresponding patient group $\cP$.
Each visit $h \in \cH$ involves a distinct set of medical codes $c \subset \mathcal{C}$ as features, where $\cC$ is the total set of medical codes appearing in $\cH$. 
In this study, $\set{C}$ contains multiple types of medical codes such as diseases, medications, procedures. 
\end{definition}

Due to the diversity among various groups of patients, there is usually a large variation in the volume of the medical codes $|\set{C}|$ across different patient groups. In this study, we consider the setting where patients are separated into two subgroups, one with basic features only and the other with both basic and extra features. 

\begin{definition}[Patients with Basic/Extra Feat.]\label{def:rich_sparse}Typically, the available data consists of a large set of EHRs $\cH$ from a wider population, which have basic patient features $\cC_b$. However, as local a medical institute collects extra features $\cC_e$, a small subset of EHRs $\cH_e \subset \cH$ further includes extra features $\cC_e$. 

\end{definition}

In this work, given the clinical record $\cH$ and both basic and extra features $\cC_b \cup \cC_e$, we aim to develop a model $g_{\theta}$ that predicts the patients' clinical outcomes $y$ so that $g_{\theta}$ can perform well on both patient groups, including those with basic features only and those with full (both basic and extra) features.


\subsection{Limitations of Traditional ML Methods}
\label{sec:prelim_limitation}
\begin{figure}[t]
\floatconts
  {fig:prelim}
  {\caption{\vspace{-2.5ex} A preliminary study with XGBoost on the two datasets.}\vspace{-3ex}}
  {%
    \subfigure[UK Biobank]{\label{fig:prelim_xgboost_ukb}%
      \includegraphics[width=0.49\linewidth]{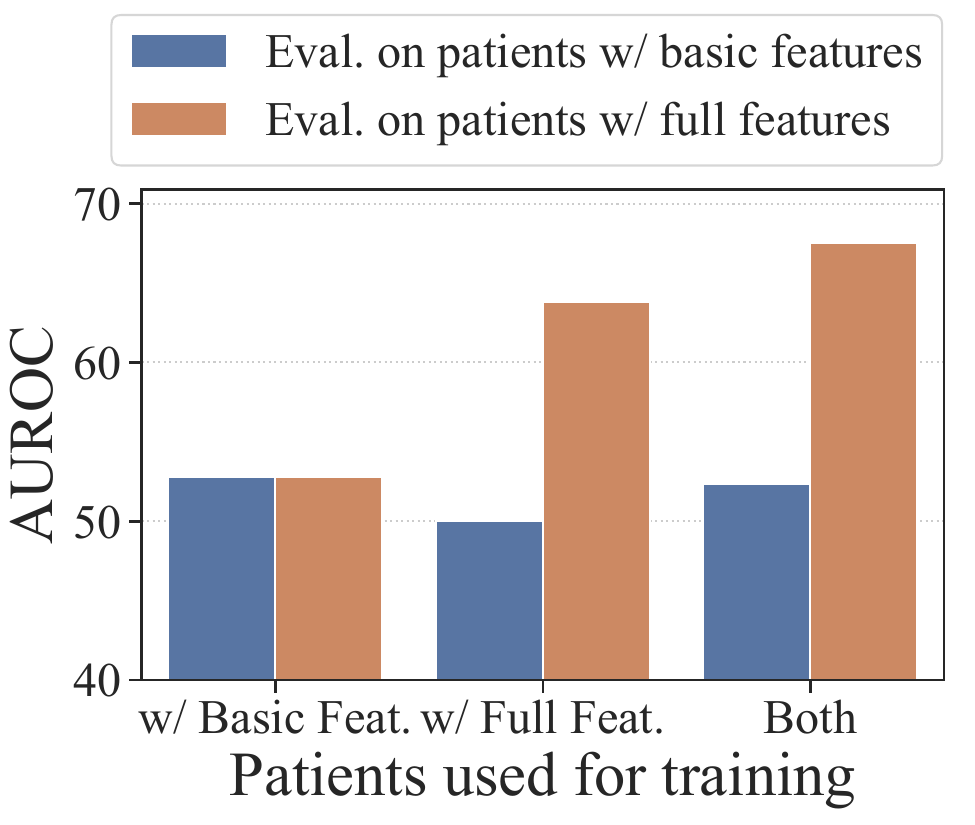}}%
    \subfigure[MIMIC-III]{\label{fig:prelim_xgboost_mimic}%
      \includegraphics[width=0.49\linewidth]{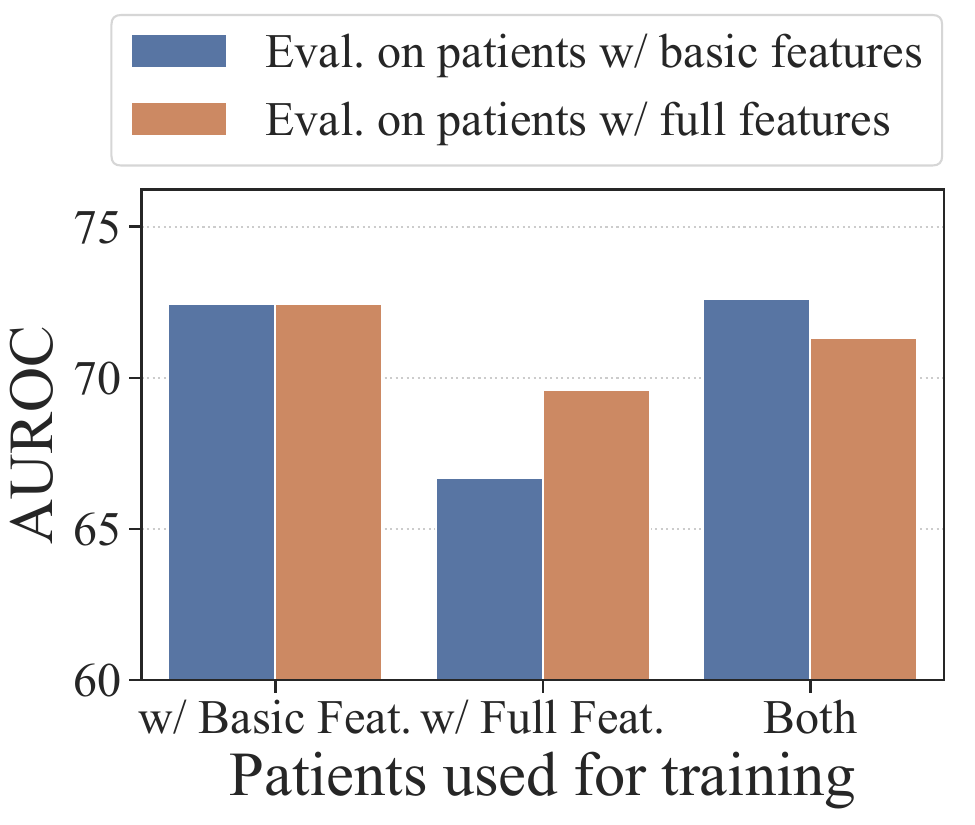}}
  }
\end{figure}

While traditional ML methods usually demonstrate strong performance in predictive EHR tasks, they encounter limitations in our specific scenario due to their inability to handle varying feature dimensions.
To illustrate this challenge, we conduct a preliminary study as shown in Figure~\ref{fig:prelim}, employing XGBoost~\citep{chen2016xgboost}, one of the most powerful ML models on two datasets: the UK Biobank~\citep{sudlow2015uk} and MIMIC-III~\citep{johnson2016mimic}.
Both datasets have a large number of patients with basic features and a smaller subset of patients with additional extra features (please refer to Section~\ref{sec:dataset} for details of statistics and task descriptions). 
We conduct three distinct sets of experiments with XGBoost-- on patients with basic features only, patients with full features only, and all patients regardless of the features they have (by filling in zeros for patients with basic features only). 
The experimental results are depicted in Figure~\ref{fig:prelim}. From the results, we have the following findings: 

\noindent \textbf{Using patients only with basic or full features hinders the model performance}:
Figure~\ref{fig:prelim} reveals that when using patients' basic or full features only, the model generally exhibits lower performance.
This is mainly due to insufficient information (when using basic features) or the limited amount of training instances (when using full features). It is necessary to design effective approaches for learning with basic and full features simultaneously.

\noindent \textbf{Simply combining patients with basic and full features yields limited performance gains}: Training with all patients, on the other hand,  involves padding the input vectors of patients with only basic features by zeros to accommodate different feature dimensions. This can potentially lead to biased information, as the model might falsely interpret these zero values as informative features. 
Therefore, incorporating all patients with all features \emph{does not necessarily enhance} the original model performance in both evaluation scenarios.

In summary, the inherent limitations of traditional ML models lead to unsatisfactory performance in the clinical setting studied in this work. These models either struggle to effectively leverage additional information or excel only in cases where patients have extra features, leaving a substantial performance gap when dealing with patients having only basic features.
This observation highlights the importance of developing effective strategies that can simultaneously incorporate patients with both basic and extra features to achieve better generalization.








\begin{figure}[t]
\floatconts
  {fig:hypergraph_illustration}
  {\caption{An illustration of the hypergraph structures.}\vspace{-1.5ex}}
  {\includegraphics[width=0.95\linewidth]{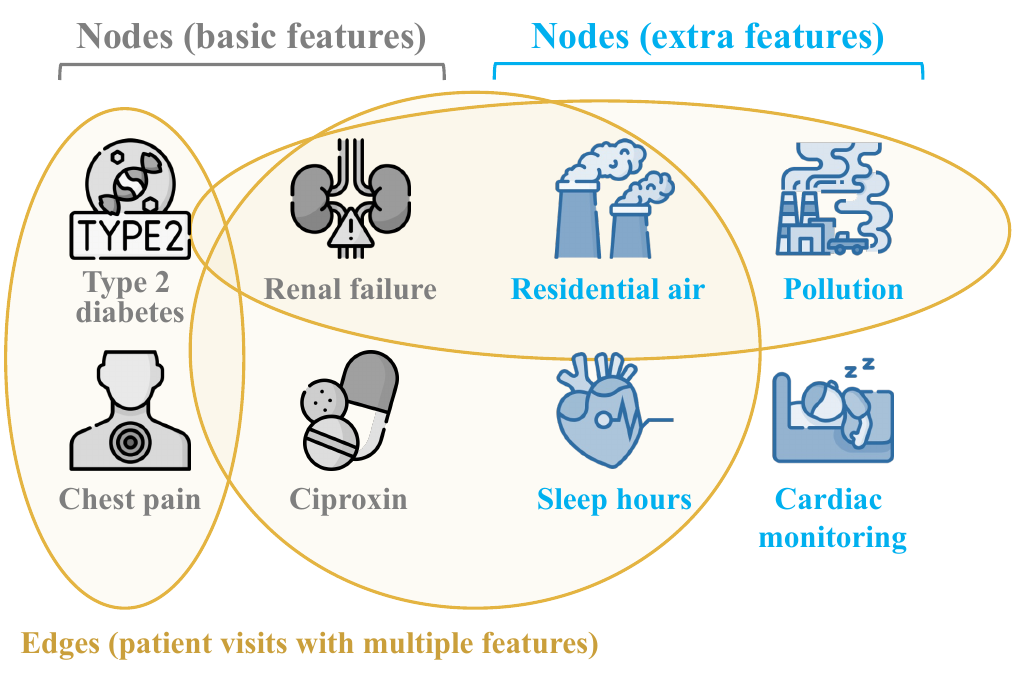}\vspace{-3ex}}
  \vspace{-3ex}
\end{figure}

\section{Method}
\label{sec:method}

\begin{figure*}[t]
\floatconts
  {fig:framework}
  {\caption{The framework of {\ours}.}}
  {\includegraphics[width=0.92\linewidth]{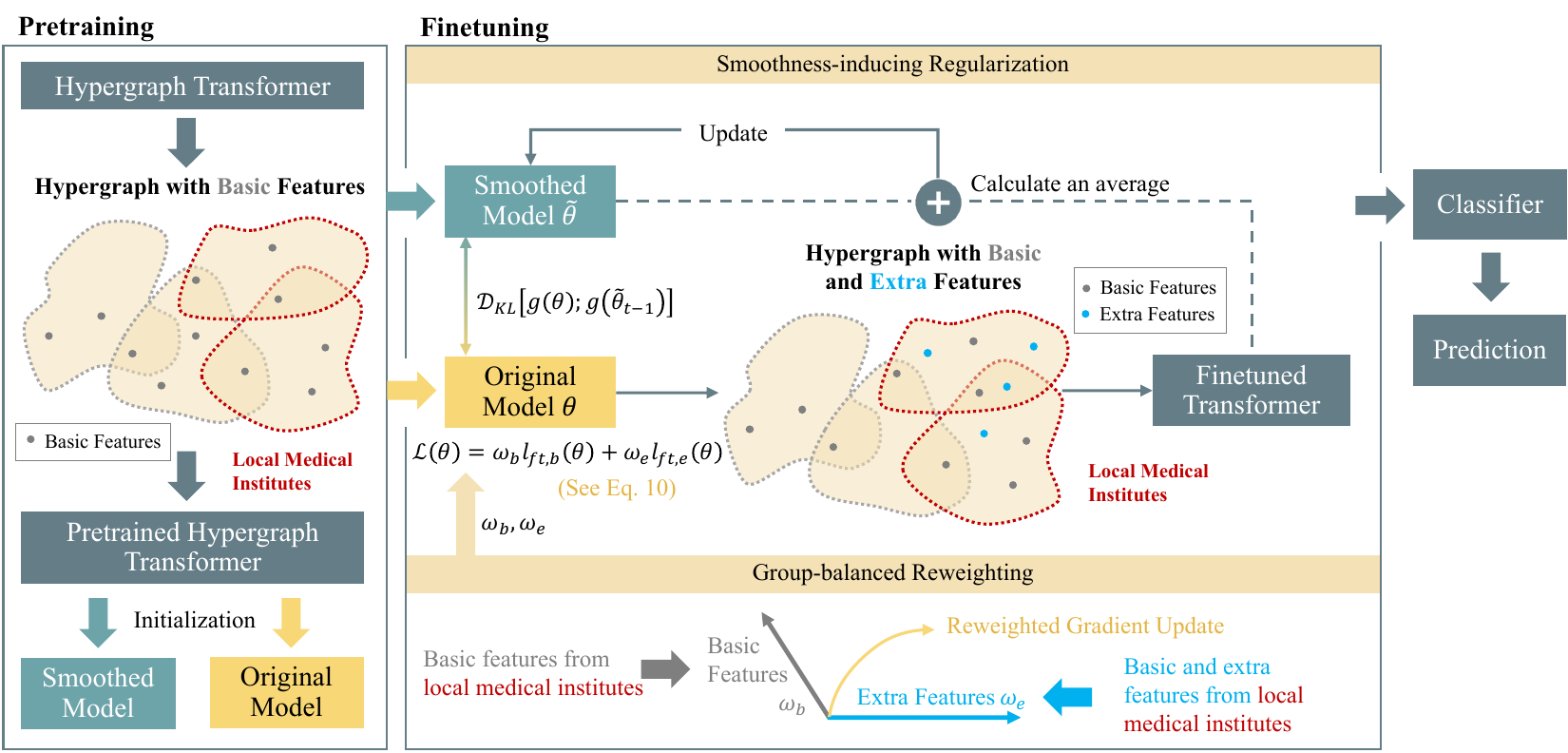}\vspace{-3ex}}
  \vspace{-1ex}
\end{figure*}

From the above analysis, it is crucial to go beyond the traditional ML modeling techniques to address the intrinsic challenges of learning with patients using basic and extra features.
Towards this end, we introduce our framework {\ours} in Figure~\ref{fig:framework}, which leverages \emph{hypergraphs} to model the EHR patient information and adopts the \emph{pretrain-then-finetune} pipeline to incorporate information from both basic and extra features. Additionally, we apply \emph{smoothness-inducing regularization} and \emph{group-balanced reweighting} techniques to mitigate issues related to catastrophic forgetting and excessive updates.

\subsection{Hypergraph Learning}

\paragraph{Graph Construction}
To better model the patient visit information as well as medical codes, it is crucial to learn the hypergraph structural information.
In this work, we model the patient visits $\cH$ as hyperedges $\cE$ and the full collection of medical codes (\ie features) $\cC_b \cup \cC_e$ as nodes $\cV$.
Each hyperedge $e \in \cE$ represents a patient visit and can connect to various nodes, where each node $v \in \cV$ stands for a medical code.
We construct $\cG_b=(\cV_b, \cE)$ as the hypergraph that includes all the patients and their basic features, and $\cG_e=(\cV, \cE_e)$ as the hypergraph that contains patients with extra features and all their features.


{Figure~\ref{fig:hypergraph_illustration} illustrates the hypergraph structures used in our approach. In this figure, the yellow circles represent hyperedges, which correspond to patient visits. Each hyperedge encompasses all the nodes (i.e., features) that are present in that particular hospital visit. This modeling approach captures the higher-order interactions among patient visits and features. Additionally, extra features can be easily incorporated into the hyperedges without the need to create new edges, providing flexibility in feature integration.}

\paragraph{Hypergraph Transformer Architecture}

Denote the representation of nodes and hyperedges on $l$-th layer as $\bX^{(l)} \in \mathbb{R}^{|\cV| \times d}$,  $\bE^{(l)} \in \mathbb{R}^{|\cE| \times d^{\prime}}$ where $d$ and $d^{\prime}$ are two hyperparameters. 
Let $\cV_{e, \bX}=\left\{\bX_{v,:}: v \in e\right\}$ denote the set of hidden representations for \emph{nodes} in the hyperedge $e$ and $\cE_{v, \bE}=\left\{\bE_{e,:}: v \in e\right\}$ denote the set of hidden representations of \emph{hyperedges} that contain the node $v$, respectively. 
In this work, we leverage the hypergraph transformer architecture $g(\cdot; \cG, \theta)$~\citep{pmlr-v193-xu22a}, which comprises several sequential layers. In the $l$-th layer, the message passing follows two steps:
\begin{align}
\setlength{\abovedisplayskip}{0.1pt}
\setlength{\belowdisplayskip}{0.1pt}
\bE_{e}^{(l)} &= f_{\mathcal{V} \rightarrow \mathcal{E}}\left(\cV_{e, \bX^{(l-1)}}\right), \label{eq:edge_embed}
\\
\bX_{v}^{(l)} &= f_{\mathcal{E} \rightarrow \mathcal{V}}\left(\cE_{v, \bE^{(l)}}\right). 
\end{align} 

To realize the propagation function $f(\cdot)$ for each layer, we use  two sub-layers: a multi-head self-attention (MHA) and a fully connected feed-forward neural network (FFNN).  
The details of these two components are deferred to Appendix~\ref{apd:hypergraph_transformer}. 
Formally, the propagation rule $f(\cdot)$ can be expressed as 
\begin{align}
\bY &= \operatorname{MHA}(\bX),\\
f(\bX) &= \operatorname{LN}(\bY + \operatorname{FFNN}(\bY)).
\end{align}

By harnessing the strong representative power of self-attention, we can identify the most relevant elements within the set for message passing, which is crucial for encoding the relationships among rich and sparse features and facilitating knowledge transfer. 

\paragraph{Predictions on Target Tasks}
To support downstream clinical prediction tasks with the learned patient representations, 
we stack a classification layer on top of the visit embeddings from \emph{all layers} $\tilde{\bE}_i^{(l)} (1\leq l \leq L)$ from Eq.~\ref{eq:edge_embed} to obtain final predictions. Specifically, for patient $i$, the prediction can be expressed as 
\begin{equation}
\hat{y}_i = g(e_i; \cG, \theta) = \sigma\left(\bW_{\operatorname{cls}}\left(\|_{l=1}^{L}\tilde{\bE}_i^{(l)}\right)\right); \label{eq:gcn_pred}
\end{equation}
where $e_i$ is the corresponding hyperedge for patient $i$, $\bW_{\operatorname{cls}}$ is a linear classification head that converts the vector to a value for binary classification and $\sigma(x) ={1}/\left({1+\exp(-x)}\right)$ is the sigmoid function. 
The target prediction task involves a binary classification task, we use the binary cross-entropy as the learning objective defined as 
\begin{equation}
\ell_{\text{cls}}(e_i, y_i) = -y\log(\hat{y}_i)-(1-y)\log(1-\hat{y}_i).
\label{eq:ce_loss}
\end{equation}

\subsection{Pretrain-then-Finetune Pipeline}
The previous section mainly discusses the hypergraph construction and learning models. 
Now, the key challenge becomes how to design an effective training scheme to better model the basic and extra clinical features of different groups of patients. 
As shown in Section~\ref{sec:prelim_limitation}, using only part of the features, as well as simply combining patients with basic and extra clinical features results in unsatisfactory performances.

To tackle this issue, we introduce a two-stage training approach, beginning with a pre-training phase followed by fine-tuning. 
Note that such a pretrain-then-finetune pipeline has been widely adopted for various domains including computer vision~\citep{chen2020simple}, text~\citep{devlin2019bert}, and time series~\citep{mcdermott2021comprehensive}. 
Initially, the predictor is trained on the hypergraph $\cG_b$. Then, to tailor the learned model  $g(\cdot; \cG_b, \theta)$ for the specific task involving patients with additional features, we fine-tune it on the hypergraph $\cG_e$. The details of this two-stage training pipeline is described in the following sections.

\subsubsection{Pretraining on Basic Features From Broader Population}

In order to equip the information from the basic features, we pretrain the model on the hypergraph $\cG_b$ where the basic features from all patients are considered. During this pretraining stage, the learning objective is denoted as
\begin{equation}
\ell_{\text{pt}}(\theta)= \mathbb{E}_{e_i\sim \cE} \ell_{\text{cls}}(e_i, y_i),
\label{eq:pt_loss}
\end{equation}
where $\ell_{\text{cls}}$ is defined in Eq.~\ref{eq:ce_loss}. This pretrained hypergraph transformer serves as the starting point in the fine-tuning stage.

\begin{algorithm2e}[t]
\small
\caption{Training Process of \ours{}.}
\label{alg:net}
\KwIn{Patient Visit $\cH = ((\cC_b, \cC_e), \cP)$, 
Numbers of iterations for pretraining and finetuning $\operatorname{Iter}_{\operatorname{pt}}$, $\operatorname{Iter}_{\operatorname{ft}}$.}
\KwOut{Finetuned hypergraph transformer $g_{\theta}$.}
\textcolor{blue}{\emph{// Step 1: Hypergraph Construction}} \\
$\cE \leftarrow \cH$, $\cV_b \leftarrow \cC_b$, $\cG_b \leftarrow (\cV_b, \cE)$ \\
\textcolor{blue}{\emph{// Step 2: Hypergraph Transformer Pretraining}}  \\
\For{$i\leftarrow 1$ \KwTo $\operatorname{Iter}_{\operatorname{pt}}$}{
  Update hypergraph transformer $g(\cdot; \cG_b, \theta)$ with $\ell_{\operatorname{pt}}(\theta)$ in Eq.~\ref{eq:pt_loss}.
}

\textcolor{blue}{\emph{// Step 3: Hypergraph Transformer Finetuning}}  \\
$\cE_b \leftarrow \cH_b$, $\cV \leftarrow (\cC_b,\cC_e)$, $\cG_e \leftarrow (\cV, \cE_e)$, $g(\cdot; \cG_e, \theta) \leftarrow g(\cdot; \cG_b, \theta)$  \\
\For{$i\leftarrow 1$ \KwTo $\operatorname{Iter}_{\operatorname{ft}}$}{
  Calculate loss for patients with basic and extra features $\ell_{\operatorname{ft,b}}$, $\ell_{\operatorname{ft,e}}$ using Eq.~\ref{eq:ft_loss}. \\
  Calculate weights $(\omega_{b}, \omega_e)$ for two groups with Eq.~\ref{eq:be_weight}. \\
  Update the hypergraph transformer $g(\cdot; \cG_e, \theta)$ with Eq.~\ref{eq:overall_learning}.
}
\end{algorithm2e}

\subsubsection{Finetuning with Customized Techniques}
After pretraining on a broader population, we then finetune our model on local data with a small number of patient visits with basic and extra features. 
To better harness the knowledge from pretraining and balance the performance between patients with basic and extra features, we design two additional techniques for our scenarios, namely \emph{Smoothness-inducing Regularization} and \emph{Group-balanced Reweighting}. The details of these two models are described as follows.

\paragraph{Smoothness-inducing Regularization}
Due to the limited data from the target task, the standard fine-tuning of the hypergraph transformer model can lead to overfitting on the training instances, resulting in poor generalization to test data~\citep{ramasesh2021effect}. 

To alleviate this issue, we maintain an additional smoothed model $g(\tilde{\theta})$, initialized by the pretrained model $g(\cdot; \cG_b, \theta)$. 
In the $t$-th step, the parameter for the smoothed model $\tilde{\theta}_t$ is updated as 
\begin{equation}
\tilde{\theta}_t=(1-\beta) \theta_t+\beta \tilde{\theta}_{t-1},
\label{eq:moving_average}
\end{equation}
where $\beta$ represents the smoothing factor, creating an exponential moving average between the parameters of the original predictor $\theta$ and the smoothed model from the previous timestep.
To encourage consistency between predictions made by the original model $g(\theta)$ and the smoothed model $g(\tilde\theta)$ during fine-tuning, we add the additional consistency regularization between the original and the smoothed model to the learning objective as
\begin{equation}
\small
    \ell_{\operatorname{ft}}(\theta)   = \ell_{\operatorname{cls}}(\theta) + \mu \mathbb{E}_{e_i\sim \cE_e} \mathcal{D}_{\mathrm{KL}}\left(g(e_i; \cG_e, \theta); g(e_i; \cG_e,\widetilde{\theta}_{t-1})\right),
\label{eq:ft_loss}
\end{equation}
where $\mathcal{D}_{\mathrm{KL}}$ is the Kullback–Leibler (KL) divergence and $\mu$ is the weight for the consistent loss. This regularization strategy effectively prevents aggressive parameter updates and enhances the model's generalization capabilities for the target prediction~\citep{tarvainen2017mean,nichol2018first}.

\paragraph{Group-balanced Reweighting}
Apart from the issue of aggressive updates, an equally, if not more, important challenge for finetuning the predictor to target patient visits is the balance between patients with basic only and extra features. To address this, we propose a reweighting scheme for dynamically adjusting the weight of patients with basic and extra features (denoted as $\omega_{\operatorname{b}}$ and $\omega_{\operatorname{e}}$, respectively) during the finetuning process. Note that in the finetuning stage, only a small subset of EHRs from patients with basic features is available, aiming to better replicate the perspective of a local medical institute.
Thus, the learning objective after the reweighting stage can be written as
\begin{equation}
\cL(\theta) = \omega_{\operatorname{b}}\ell_{\operatorname{ft,b}}(\theta) + \omega_{\operatorname{e}}\ell_{\operatorname{ft,e}}(\theta).
\label{eq:overall_learning}
\end{equation}
Here $\ell_{\operatorname{ft,b}}(\theta)$ and $\ell_{\operatorname{ft,e}}(\theta)$ stand for the finetuning loss, which is defined in Eq.~\ref{eq:ft_loss}.
When learning with two groups of patients simultaneously, we hypothesize that an ideal choice of $\omega_{\operatorname{b}}$ and $\omega_{\operatorname{e}}$ would provide the biggest reduction on the training loss of the two groups. 
We approximate the loss reduction using first-order Taylor expansion:
\begin{align}
&\Delta\ell^{(t)}
= \sum_{i \in \{\operatorname{b}, \operatorname{e}\}} \left(\ell_{i}(\theta-\alpha \nabla_{\theta} \mathcal{L}(\theta)) -\ell_{i}(\theta)\right) \label{eq:grad_step} \\ 
&\approx -\alpha\sum_{i\in \{\operatorname{b}, \operatorname{e}\}}\sum_{j\in \{\operatorname{b}, \operatorname{e}\}}\omega^{(t)}_i\left(\nabla_{\theta}\ell_i(\theta)\right)^{\texttt{T}}\nabla_{\theta}\ell_j(\theta), \label{eq:app_grad_step}
\end{align}
{where $\alpha$ is the learning rate.} In addition, we avoid the potential rapid change of weights for finetuning stability, by adding a KL divergence regularization between $\boldsymbol{\omega}=(\omega_{\operatorname{b}}, \omega_{\operatorname{e}})$ at different steps. This leads to the following optimization target:
\begin{align}
&\min _{\omega^{(t)}} \ \Delta\ell^{(t)} + \mathcal{D}_{\operatorname{KL}}(\boldsymbol{\omega}^{(t)}, \boldsymbol{\omega}^{(t-1)}), \\
&\text{ s.t.} \quad \omega_{\operatorname{b}} + \omega_{\operatorname{e}}=1. \label{eq:constrain_reg}
\end{align}
By using the Lagrangian multiplier with KKT conditions, we obtain the closed-form solution for the weight on patients with basic and extra features in step $t$ as:
\begin{equation}
\omega_i^{(t)}=\frac{\omega_i^{(t-1)} \cdot \exp \left(\langle s_i, \sum_j s_j\rangle\right)}{\sum_{k\in \{\operatorname{b}, \operatorname{e}\}} \omega_k^{(t-1)} \cdot \exp \left(\langle s_k, \sum_j s_j\rangle\right)},
\label{eq:be_weight}
\end{equation}
where $i\in \{\operatorname{b}, \operatorname{e}\}$, $s_i=\nabla_{\theta}\ell_i(\theta)$ is the gradient for loss $l_i$. 
The ideal solution naturally takes into account the similarity of gradients between patients who have basic features and those with extra features. It prioritizes the allocation of weights that share more common needs with others, to enhance the robustness of the model across different patients.

\subsection{Overall Algorithm}
To better illustrate the learning procedure, the overall procedure is listed in Algorithm~\ref{alg:net}.
It is worth noting that \ours{} can be trained in an end-to-end manner, without heavy parameter tuning. 


\section{Experiments}
\label{sec:exp}

\begin{table}[t]
\floatconts
  {tab:dataset_stats}
  {\caption{Dataset Statistics.}\vspace{-1ex}}
  {
    \renewcommand\arraystretch{0.93}
  \resizebox{\linewidth}{!}{
  \begin{tabular}{lcc}
  \toprule
  \bfseries Stats & \bfseries UK Biobank & \bfseries MIMIC-III\\
  \midrule
  \# basic features & 642 & 846 \\
  \# extra features & 1371 & 6577 \\
  \# health records & 1629 & 12353\\
  └── \# train samples w/ basic only & 1140 & 8647 \\
  └── \# train samples w/ extra & 164 & 1235\\
  └── \# validate samples & 162 & 1236\\
  └── \# test samples & 488 & 3706\\
  \bottomrule
  \end{tabular}
  }
  }
  \vspace{-2ex}
\end{table}

\subsection{Datasets and Tasks}
\label{sec:dataset}

We conduct experiments on two datasets: UK Biobank~\citep{sudlow2015uk} and MIMIC-III~\citep{johnson2016mimic}, with the statistics shown in Table~\ref{tab:dataset_stats}. 

The UK Biobank dataset~\citep{sudlow2015uk} is a comprehensive biomedical national biobank and research initiative based in the United Kingdom. It involves participants aged 40 to 69 who were enrolled between 2006 and 2010. It recruits a small subset of patients to take part in an assessment, where extra features such as sleep hours and cardiac monitoring are recorded\footnote{{Some other features could be found at \url{https://biobank.ctsu.ox.ac.uk/crystal/browse.cgi}}}. We conduct an outcome prediction task which predicts whether the patients with type 2 diabetes would experience cardiovascular disease (CVD) endpoints within 10 years after their initial diagnosis. Specifically, CVD endpoints represent the presence of coronary heart disease (CHD), congestive heart failure (CHF), dilated cardiomyopathy (DCM), myocardial infarction (MI), or Stroke. Please refer to Appendix~\ref{apd:more_data} for preprocessing details.


The MIMIC-III dataset~\citep{johnson2016mimic} contains over 40,000 de-identified patients in critical care units of the Beth Israel Deaconess Medical Center from 2001 to 2012. We conduct phenotyping prediction on MIMIC-III, which is formulated as a multi-label classification on the 25 pre-defined phenotypes by~\cite{harutyunyan2019multitask}. Specifically, given the patients' health records, we aim to predict whether the 25 acute care conditions are present in their next visits. See Appendix~\ref{apd:more_data} for the detailed list of the phenotypes.
In the preprocessing stage, we extract patients with multiple hospital visits and create pairs of consecutive visits for each patient. For each pair, we extract the diseases, medications, procedures, and services in the former visit as input features. Among them, diseases are considered as the basic features, and the others are considered as extra features.
{MIMIC-III is in a simulated setting and thus the extra features are intentionally masked out for most patients, even though some of them might have that information available in the dataset.}
This is because diseases are typically readily available through claims data, and often serve as the primary focus in various analytical tasks~\citep{wu2023collecting}.
The phenotypes present in the latter visit serve as the corresponding labels.

For both datasets, we construct two subgroups for evaluation. To ensure a fair comparison, we maintain the same set of patients in both subgroups, with one group having only basic features, and the other having additional extra features. Moreover, we evaluate {\ours} and all baselines on both subgroups to show their capabilities in two different scenarios.

\setlength{\tabcolsep}{4pt}

\begin{table*}[t]
\floatconts
  {tab:main_exp}
  {\vspace{-2ex}\caption{Performance on UK Biobank and MIMIC-III compared with baselines. ``P/F'' stands for methods with pretrain-then-finetuning. {``HyG'' represents hypergraph.} \textbf{Bold} indicates the best result across all models. {The result is averaged over 5 runs. * denotes statistical significant results ($p<0.05$).}} \vspace{-2.5ex}}
  {
  \renewcommand\arraystretch{0.93}
\resizebox{0.98\linewidth}{!}{
\begin{tabular}{l|cccccc|cccccc}
\toprule
\multirow{3.5}{*}{\bfseries Model} & \multicolumn{6}{c|}{\bfseries UK Biobank} & \multicolumn{6}{c}{\bfseries MIMIC-III} \\
\cmidrule(lr){2-7} \cmidrule(lr){8-13}
& \multicolumn{3}{c}{\bfseries Basic} &  \multicolumn{3}{c|}{\bfseries Full} & \multicolumn{3}{c}{\bfseries Basic} &  \multicolumn{3}{c}{\bfseries Full}           \\ 
\cmidrule(lr){2-4} \cmidrule(lr){5-7} \cmidrule(lr){8-10} \cmidrule(lr){11-13}
&  ACC &  AUROC &  AUPR & ACC &  AUROC &  AUPR &  ACC &  AUROC &  AUPR & ACC &  AUROC &  AUPR \\
\midrule
LR w/ Basic Feat. & 67.90 & 51.76 & 46.69 & 67.90 & 51.76 & 46.69 & 75.85 & 72.31 & 54.25 & 75.85 & 72.31 & 54.25 \\
LR w/ Full Feat. & 64.20 & 57.36 & 48.05 & 69.14 & 68.50 & 64.32 & 74.98 & 66.86 & 50.30 & 74.59 & 67.41 & 49.11 \\
LR w/ Both & 62.96 & 52.35 & 43.58 & 67.90 & 64.79 & 60.82 & 75.01 & 72.28 & 54.25 & 75.10 & 68.09 & 49.50 \\
XGBoost w/ Basic Feat. & 61.73 & 52.82 & 44.54 & 61.73 & 52.82 & 44.54 & 75.97 & 72.44 & 54.61 & 75.97 & 72.44 & 54.61\\
XGBoost w/ Full Feat. & 64.20 & 50.00 & 35.80 & 64.20 & 63.79 & 57.27 & 76.83 & 66.71 & 50.19 & 76.06 & 69.61 & 51.85 \\
XGBoost w/ Both & 64.20 & 52.35 & 44.84 & 64.20 & 67.51 & 56.63 & 76.88 & 72.60 & 54.54 & 75.31 & 71.33 & 53.25 \\
Transformer w/ Basic Feat. & 62.96\scriptsize±0.32 & 59.88\scriptsize±0.21 & 46.42\scriptsize±0.32 & 62.96\scriptsize±0.32 & 59.88\scriptsize±0.21 & 46.42\scriptsize±0.32 & 72.45\scriptsize±0.92 & 74.64\scriptsize±0.61 & 59.47\scriptsize±0.76 & 72.45\scriptsize±0.92 & 74.64\scriptsize±0.61 & 59.47\scriptsize±0.76 \\
Transformer w/ Full Feat. & 64.20\scriptsize±0.00 & 41.35\scriptsize±0.28 & 31.43\scriptsize±0.45 & 64.20\scriptsize±0.00 & 54.21\scriptsize±0.35 & 37.24\scriptsize±0.42 & 71.72\scriptsize±0.13 & 72.76\scriptsize±0.24 & 57.46\scriptsize±0.36 & 71.72\scriptsize±0.13 & 72.92\scriptsize±0.25 & 57.45\scriptsize±0.50 \\
Transformer w/ Both & 64.20\scriptsize±0.00 & 40.22\scriptsize±0.36 & 30.71\scriptsize±0.40 & 64.20\scriptsize±0.00 & 54.31\scriptsize±0.39 & 37.63\scriptsize±0.40 & 72.21\scriptsize±0.27 & 74.55\scriptsize±0.42 & 59.49\scriptsize±0.69 & 71.47\scriptsize±0.26 & 72.35\scriptsize±0.47 & 56.07\scriptsize±0.59 \\
\midrule
HyG + vanilla P/F & 62.96\scriptsize±0.92 & 56.79\scriptsize±0.38 & 48.74\scriptsize±0.26 & \textbf{69.14\scriptsize±0.99} & 71.56\scriptsize±0.37 & 61.26\scriptsize±1.84 & 75.04\scriptsize±0.69 & 75.77\scriptsize±0.59 & 62.66\scriptsize±0.38 & 74.67\scriptsize±0.60 & 76.54\scriptsize±0.92 & 63.36\scriptsize±0.84 \\
\midrule
HyG + Reweight P/F & 65.43\scriptsize±0.00 & 57.33\scriptsize±0.49 & 49.40\scriptsize±0.09 & 65.43\scriptsize±0.00 & 70.23\scriptsize±0.58 & 64.44\scriptsize±0.20 & 75.10\scriptsize±1.14 & 76.04\scriptsize±0.53 & 63.14\scriptsize±0.96 & 74.36\scriptsize±1.29 & 76.65\scriptsize±0.87 & 62.72\scriptsize±0.80 \\
HyG + AUX-TS P/F & 62.96\scriptsize±0.83 & 56.13\scriptsize±0.44 & 49.63\scriptsize±0.15 & 64.20\scriptsize±0.86 & 58.69\scriptsize±0.43 & 38.09\scriptsize±1.36 & 75.89\scriptsize±1.01 & 77.06\scriptsize±0.40 & 66.27\scriptsize±0.62 & 64.69\scriptsize±1.27 & 63.84\scriptsize±1.34 & 47.02\scriptsize±0.97 \\
HyG + G-Adv P/F & 65.43\scriptsize±0.00 & 59.98\scriptsize±0.27 & 48.50\scriptsize±0.11 & 64.20\scriptsize±1.12 & 57.29\scriptsize±0.78 & 39.78\scriptsize±2.01 & 73.79\scriptsize±1.20 & 76.71\scriptsize±0.58 & 62.56\scriptsize±0.94 & 73.15\scriptsize±1.35 & 75.23\scriptsize±0.97 & 60.85\scriptsize±0.73 \\
HyG + ForkMerge P/F & 64.20\scriptsize±1.10 & 58.39\scriptsize±0.35 & 50.28\scriptsize±0.21 & 66.67\scriptsize±0.94 & 74.20\scriptsize±0.23 & 64.28\scriptsize±0.19 & 75.56\scriptsize±1.70 & 76.11\scriptsize±0.54 & 64.14\scriptsize±0.88 & 70.85\scriptsize±1.21 & 70.12\scriptsize±1.63 & 54.75\scriptsize±1.18 \\
\midrule
\rowcolor{teal!10} \makecell[l]{{\ours} \\ \hphantom{2pt}(HyG + proposed P/F)}
& \textbf{72.84\scriptsize±0.71}* & \textbf{60.11\scriptsize±0.43} & \textbf{50.78\scriptsize±0.16}* & 67.90\scriptsize±0.88 & \textbf{74.40\scriptsize±0.28} & \textbf{65.54\scriptsize±0.12}* & \textbf{78.17\scriptsize±1.09}* & \textbf{81.00\scriptsize±0.42}* & \textbf{69.54\scriptsize±0.70}* & \textbf{77.06\scriptsize±0.69}* & \textbf{79.56\scriptsize±0.92}* & \textbf{67.13\scriptsize±0.55}* \\
\bottomrule
\end{tabular}
}

  }
\end{table*}

\subsection{Baselines}
We mainly compare \ours{} with two groups of baselines. The detailed description of baselines is deferred to Appendix \ref{apd:baselines}.{We employ Accuracy, Macro AUROC and Macro AUPR as evaluation metrics.}
\paragraph{Traditional ML Baselines}
These methods do not leverage graph structure to model the relationships between patients and features. For these methods, we consider three variants --- \emph{basic} (patients with basic features only), \emph{extra} (patients with extra features only), and \emph{combined} (considering all patients, and zero-padding is used to ensure the alignment of basic and extra features dimensions). 
In this group of baselines, we consider three techniques: (1) \textbf{Logistic Regression}~(LR,~\citet{keyhani2008electronic}),   (2) \textbf{XGBoost}~\citep{chen2016xgboost} and  (3) \textbf{Transformer}~\citep{li2020behrt}. 

\paragraph{Pretrain-then-Finetune Baselines} These baselines propose additional training techniques to facilitate knowledge transfer and improve the model's generalization ability. Specifically, we consider the following  baselines: (4) \textbf{PT-FT}~\citep{pmlr-v193-xu22a}, (5) \textbf{Reweight}~\citep{li2023multi}, (6) \textbf{AUX-TS} \citep{han2021adaptive}, (7) \textbf{G-Adv} \citep{dai2022graph,liu2021subtype}, (8) \textbf{ForkMerge} \citep{jiang2023forkmerge}.
We are aware that there are additional techniques for EHR-based clinical predictions, however, they either focus on designing neural architectures~\citep{zhu2021variationally} or leverage additional knowledge~\citep{pmlr-v174-park22a,cui2023survey,jiang2023graphcare,xu2024ram}, thus are orthogonal to the focus of this work. 

\subsection{Implementation Details}
We implement our model in PyTorch~\citep{paszke2019pytorch}. 
We {tune the learning rate ($\alpha$) in the range of \{1e-4, 2e-4, 1e-3\} and} set it as 1e-3 in the pretraining stage and 2e-4 in the finetuning stage.
We use Adam~\citep{kingma2014adam} as the optimizer with a weight decay of 1e-3. We set $\mu$ in Eq.~\ref{eq:ft_loss} as 0.5, $\beta$ in Eq.~\ref{eq:moving_average} as 0.5, and number of layers $l$ in the hypergraph transformer as 3. 
For the local 
The experiment is run on a single NVIDIA Titan RTX GPU.
We study the effect of $\mu$ and $\beta$ in Section~\ref{sec:parameter}.

\subsection{Experimental Results}


Table~\ref{tab:main_exp} summarizes the experimental results of {\ours} compared with baselines. Note that \emph{AUROC} is the main metric for the model performance. From the results, we have the following findings:

\noindent $\diamond$ Models following the pretrain-then-finetune pipeline generally exhibit better performance compared to traditional ML methods which face challenges in implementing this pipeline due to the dimension mismatch issue mentioned in section \ref{sec:prelim_limitation}.

\noindent $\diamond$ Directly leveraging the vanilla pretrain-then-finetune can be suboptimal, as it performs well on patients with full features but is less satisfactory for patients with basic features. 
We attribute this phenomenon to the issue of \emph{catastrophic forgetting}, where the model may \emph{forget} the knowledge learned during the pretraining stage~\citep{mehta2023empirical}. This further highlights the need for designing effective fine-tuning techniques to circumvent this issue.

\noindent $\diamond$ When compared to other transfer learning techniques, our framework achieves better performance. 
This is because some baselines (e.g. G-Adv, ForkMerge) are mainly proposed to improve the robustness of finetuning without modeling the relations between patients with basic and extra features, while other baselines (e.g. Reweight, AUX-TS) mainly use loss scales and gradient cosine similarity to reweight different group, which fail to consider overall loss reduction. On the contrary, we propose dynamically reweight different patient groups to avoid sacrificing the average performance.

\begin{figure}[t]
\floatconts
  {fig:ablation}
  {\caption{\vspace{-1ex}Effect of different components of {\ours} on the two datasets. SR and GR stands for Smoothness-inducing Regularization and Group-balanced Reweighting, respectively.}\vspace{-2.5ex}}
  {\vspace{-1ex}\includegraphics[width=0.93\linewidth]{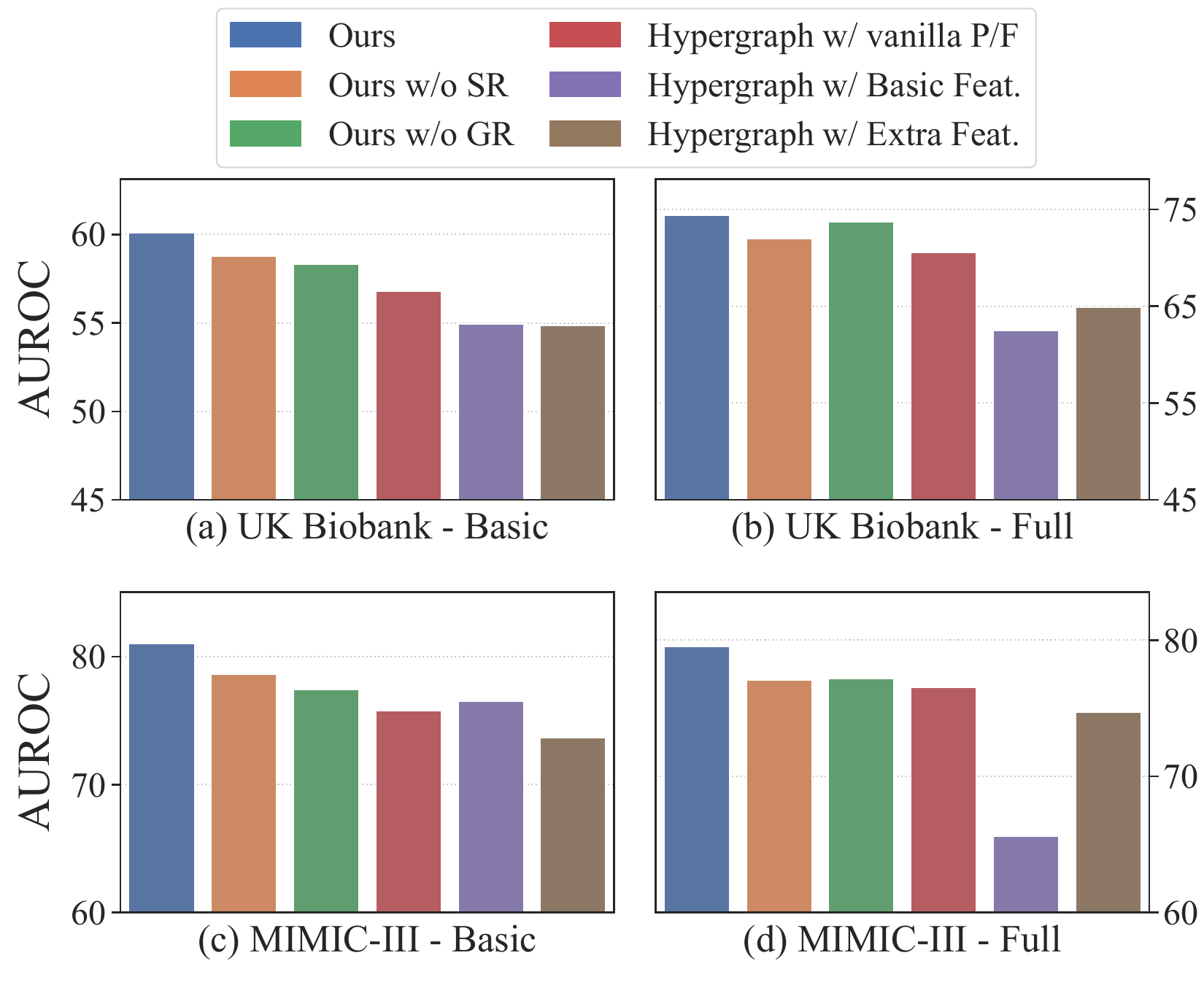}\vspace{-3ex}}
  \vspace{-1.5ex}
\end{figure}

\subsection{Ablation Study}
\label{sec:ablation}

We study the effect of different components of {\ours} on the two datasets, shown in Figure~\ref{fig:ablation}. We observe that both Smoothness-inducing Regularization and Group-balanced Reweighting are beneficial to the model performance, as they address the catastrophic forgetting issue and identify an optimized gradient direction that balances between basic features and extra features. Additionally, we also demonstrate that the pretrain-then-finetune pipeline generally enhances the model performance in both evaluation scenarios. Simply using only patients with basic features, or patients with extra features do not harness all the information.

\subsection{Parameter Study}
\label{sec:parameter}

\begin{figure}[t]
\floatconts
  {fig:parameter}
  {\vspace{-1ex}\caption{Parameter studies on UK Biobank.}\vspace{-2ex}}
  {\vspace{-1ex}
    \subfigure[$\beta$]{\label{fig:para_beta}%
      \includegraphics[width=0.48\linewidth]{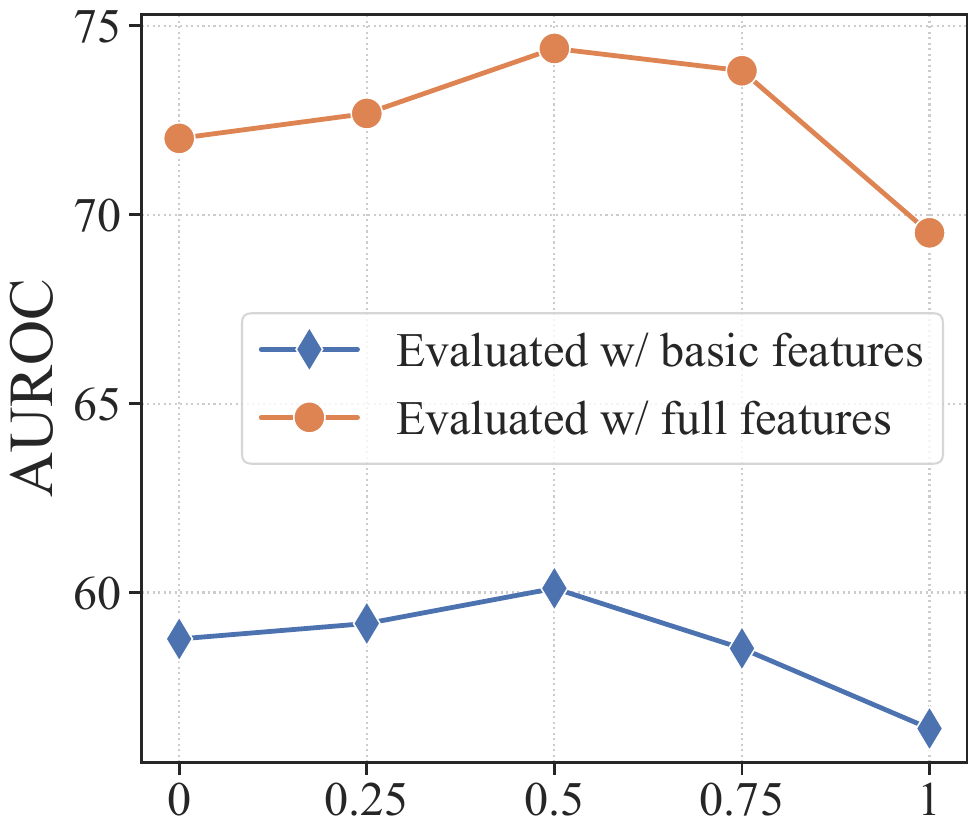}}%
    \subfigure[$\mu$]{\label{fig:para_mu}%
      \includegraphics[width=0.48\linewidth]{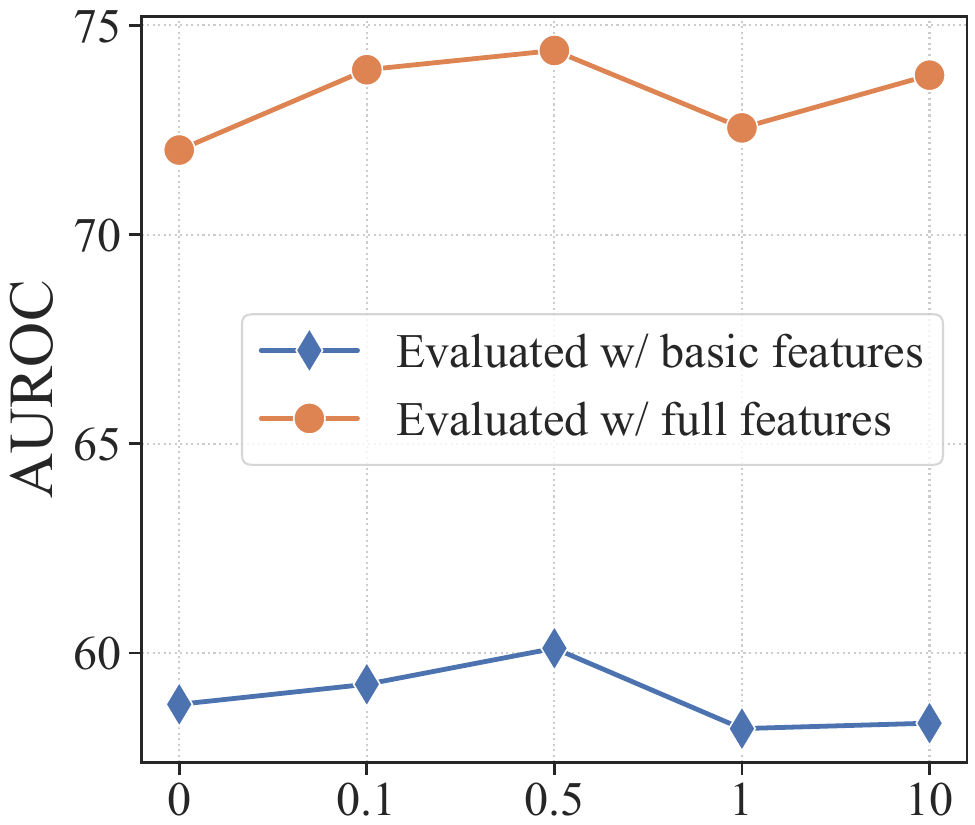}}
    \vspace{-2ex}
  }
\end{figure}

We study the effect of $\beta$ and $\mu$ in Eq.~\ref{eq:moving_average} and Eq.~\ref{eq:ft_loss}, respective, in Figure~\ref{fig:parameter}. 
The results indicate that the model achieves its optimal performance when the smoothing factor $\beta$ is set to 0.5, which evenly balances the influence of both the smoothed model and the original model. When $\beta$ equals 1, only the smoothed model is considered, while $\beta$ at 0 implies the model operates without smoothness-inducing regularization. In both extreme cases, the model neglects information from either the smoothed or original model, leading to reduced performance.
Additionally, the parameter $\mu$ serves as the weight for the consistency loss in Eq.~\ref{eq:ft_loss}. A higher value of $\mu$ signifies a greater alignment with the previous smoothed model, while a lower value of $\mu$ places more emphasis on the original model. The optimal model performance is achieved when $\mu$ is set to 0.5.

\subsection{Study on Different Balancing Methods}
\begin{figure}[t]
\floatconts
  {fig:diff_gradient}
  {\caption{\vspace{-1ex}Performance of {\ours} with different balancing method.}\vspace{-2.5ex}}
  {\includegraphics[width=0.95\linewidth]{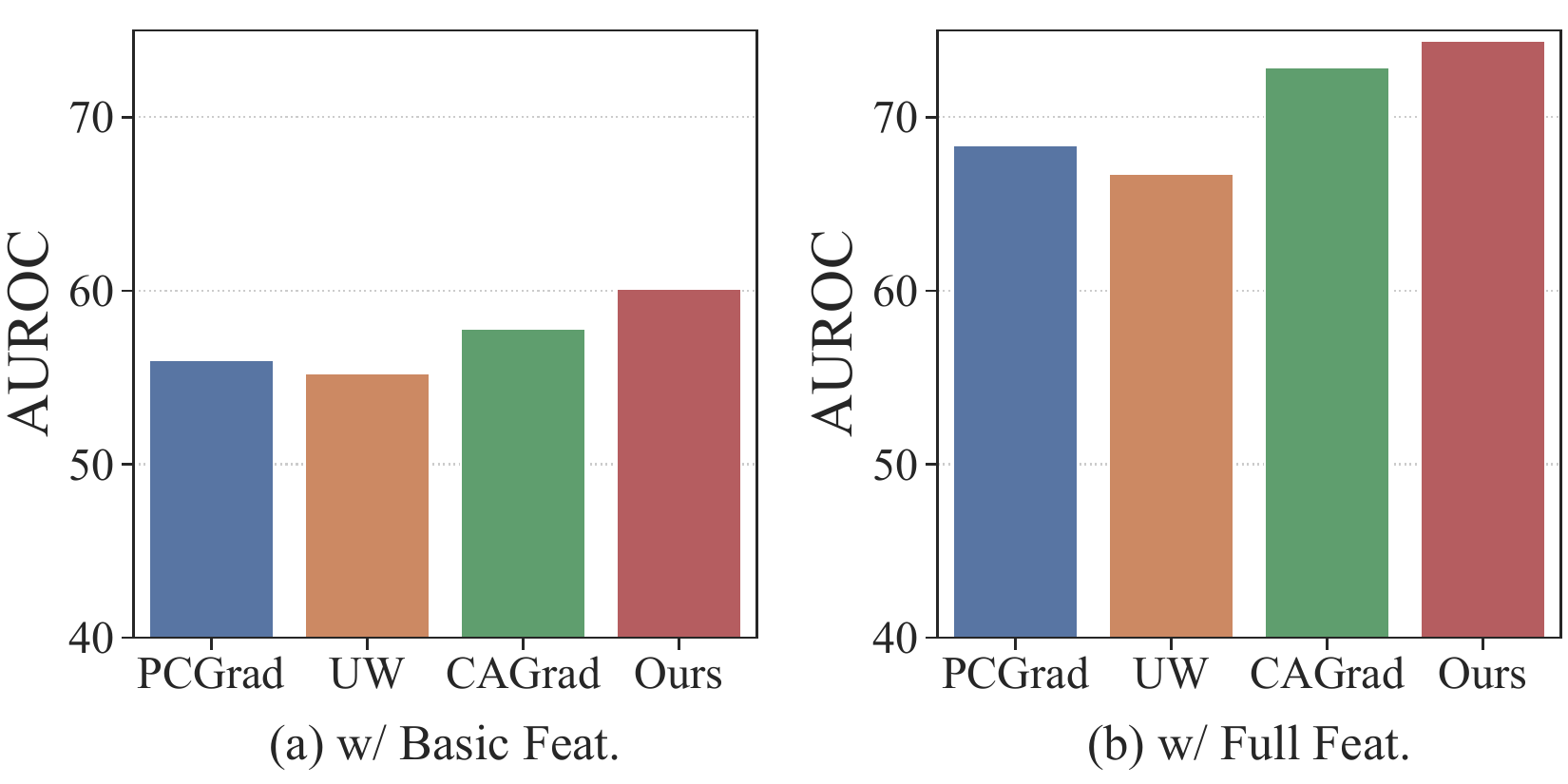}\vspace{-3ex}}
  \vspace{-1ex}
\end{figure}
As {\ours} includes a reweighting step to balance the weight between two different patient groups, we further compare with other generic reweighing methods originally proposed for multi-task learning to understand the benefit of our design further.  
Specifically, we compare with three representative methods: \emph{Uncertainty Weighting}~\citep{uw} that leverages task homoscedastic uncertainty to weight each group; \emph{CAGrad}~\citep{cagrad} and \emph{PCGrad}~\citep{pcgrad}, which design gradient harmonization approaches to avoid negative transfer.

Figure~\ref{fig:diff_gradient} illustrates the result. We observe that UW does not perform well in our setting, as we observe that the training process can be highly unstable, especially for patients with full features. Incorporating gradient harmonization approaches is beneficial, but the gain is not so significant as they do not take the overall loss reduction into account. These results corroborate the advantage of our proposed group-balanced reweighting.

\subsection{A Closer Look at the Finetuning Stage}
\begin{figure}[t]
\floatconts
  {fig:learning_curve}
  {\caption{\vspace{-3ex}Learning curve of the finetuning stage from {\ours}, compared with hypergraph transformer with vanilla pretraining and finetuning on UK Biobank.}\vspace{-3ex}}
  {%
    \subfigure[HG + vanilla P/F]{\label{fig:learning_curve_hypehr_pf}%
      \includegraphics[width=0.48\linewidth]{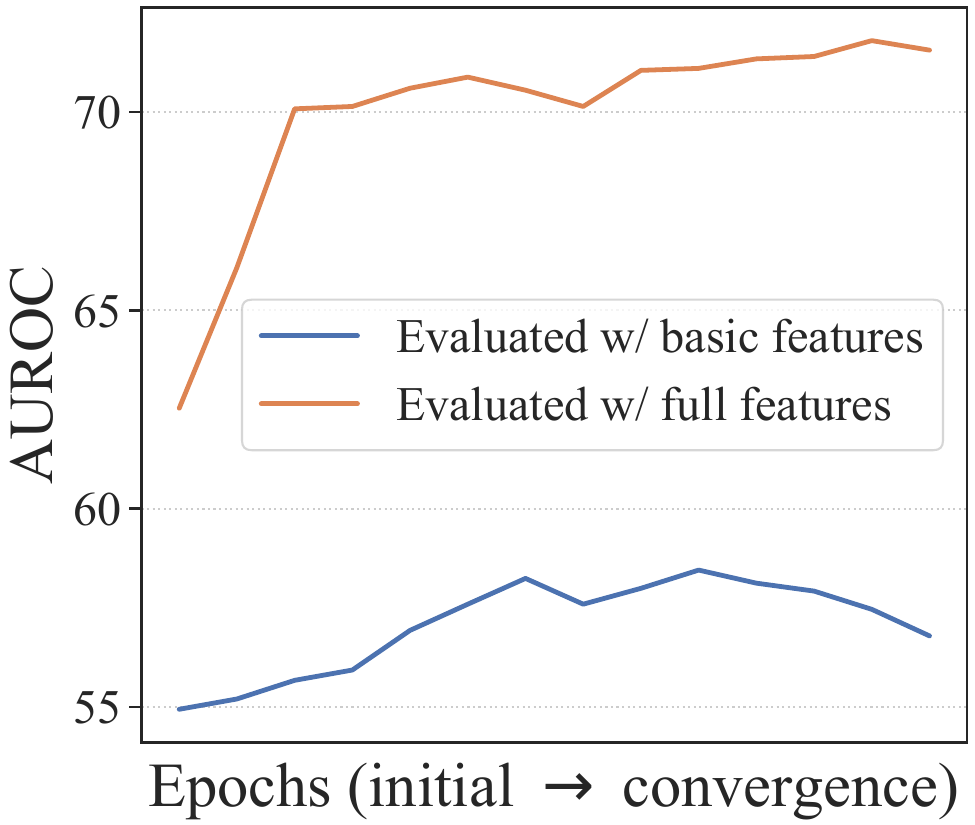}}
    \subfigure[{\ours}]{\label{fig:learning_curve_ours}%
      \includegraphics[width=0.48\linewidth]{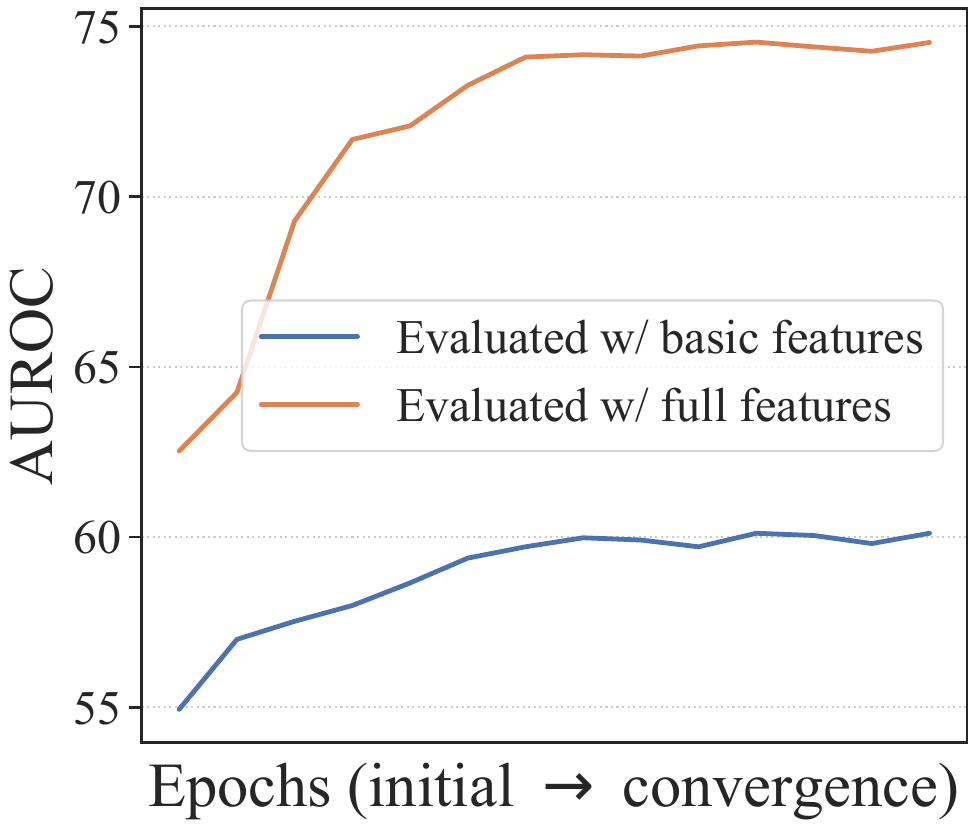}}%
  }
  \vspace{-1.5ex}
\end{figure}

Figure~\ref{fig:learning_curve} illustrates the learning curve for the finetuning phase of {\ours}, in comparison with the hypergraph transformer incorporating vanilla pretraining and finetuning. Our model effectively balances performance across patients with basic and extra features, enhancing their performance simultaneously. In contrast, the baseline model experiences a decline in performance on patients with basic features while showing continuous improvement for those with full features. This disparity occurs because the baseline model fails to maintain a balance between the two evaluation scenarios and gradually forgets pretraining information during finetuning.



\vspace{-1ex}
\section{Conclusion}
\vspace{-1ex}

We introduce \ours{}, a framework leveraging hypergraph structures within a pretrain-then-finetune framework for EHR modeling, facilitating seamless integration of additional features. 
Additionally, we propose two techniques:
(1) \emph{Smoothness-inducing Regularization} and
(2) \emph{Group-balanced Reweighting},
to enhance model robustness during finetuning.
Through experiments on two real EHR datasets, we demonstrate that \ours{} consistently outperforms various baselines, maintaining a balance between patients with basic and extra features.

\vspace{-1ex}
\section{Limitation}
\vspace{-1ex}

In this work, we mainly focus on other medical codes as extra features, but in real clinical applications, there could be other types of features from other modalities, e.g. text~\citep{pmlr-v174-park22a}, images~\citep{lee2023learning}, or time series~\citep{mcdermott2021comprehensive}. 
It is important to design techniques to incorporate data from these modalities to further broaden the application range of \ours{}.

Besides, the inclusion of a pretrain-then-finetune pipeline leads to longer training time, which can be problematic when there are large amount of patient data. A promising avenue for future research involves designing efficient training techniques to improve the scalability of our proposed framework.

\section*{Acknowledgement}
We thank the anonymous reviewers and area chairs for valuable feedbacks. 
This research was partially supported by the Emory Global Diabetes Center of the Woodruff Sciences Center, Emory University. Research reported in this publication was supported by the National Institute Of Diabetes And Digestive And Kidney Diseases of the National Institutes of Health under Award Number K25DK135913. 
The research also receives partial support by the National Science Foundation under Award Number IIS-2145411. The content is solely the responsibility of the authors and does not necessarily represent the official views of the National Institutes of Health.

\bibliography{chil-sample}

\appendix




\section{Details about Hypergraph Transformer}
\label{apd:hypergraph_transformer}
MHA computes the attention in parallel $h$ heads as: 
\begin{align}
& \operatorname{MHA}(\bX)=\operatorname{Concat}\left(\text{head}_1, \ldots, \text {head}_h\right) \bW_o, \\
& \text{head}_i=\operatorname{Softmax}\left(\bW_{q_i}(\bX \bW_{k_i})^T / \sqrt{d_h}\right) \bX \bW_{v_i},
\end{align}

where $\bX \in \mathbb{R}^{n\times d}$ is the input, $d_h = d / h$, and $\bW_{q_i} \in \mathbb{R}^{1 \times d_h}$, $\bW_{k_i}, \bW_{v_i} \in \mathbb{R}^{d \times d_h}$ are query, key, and value projection matrices for $i$-th head, respectively\footnote{Here the size of $\bW_{q_i}$ is $\mathbb{R}^{1 \times d_h}$ since we only need to generate an aggregated embedding for each node/hyperedge.}. $\bW_o \in \mathbb{R}^{d \times d}$ is an output projection matrix. 
The fully connected feed-forward neural network (FFNN) comprises two linear layers with an activation function: 
\begin{equation}
\operatorname{FFNN}(\bX)=\sigma\left(\bX \bW_{f_1}+\bb_1\right) \bW_{f_2}+\bb_2
\end{equation} 
where $\bW_{f_1} \in \mathbb{R}^{d \times d_m}, \bW_{f_2} \in \mathbb{R}^{d_m \times d}$, $\sigma(\cdot)$ is the activate function. A residual connection is used followed by a layer normalization.

\section{Datasets and Tasks Details}
\label{apd:more_data}

\paragraph{UK Biobank}
In the preprocessing stage, we consider
We consider patients in both inpatient and outpatient EHR with type 2 diabetes (ICD10 code of ‘E11.XX’). The task labels are whether the patients develop CHD (ICD10 code of ‘I25.XX’), CHF (ICD10 code of ‘I50.XX’), DCM (ICD10 code of ‘I42.XX’), MI (ICD10 code of ‘I21.XX’) and Stroke (ICD10 code of ‘I66.XX’) within 10 years of the diagnosis of type 2 diabetes. 
Note that we also consider death with causes of CHD, CHF, MI or Stroke as a positive outcome (label should be 1).
Patients with an interval between their initial and last medical record of less than 10 years or with a documented medical history of any of the specified outcomes (CHD, CHF, DCM, MI, or Stroke) before their initial diabetes diagnosis are excluded from our analysis.
To formulate a subset of the patients with extra features, we extract those patients who have been enrolled in the UKB assessment within two years before their initial diabetes diagnosis, as the assessment introduces additional features.

\paragraph{MIMIC-III.}
Table~\ref{tab:mimic_phenotypes} presents a detailed list of the 25 pre-defined phenotypes, which are identified using Clinical Classifications Software
(CCS) from the Healthcare Cost and Utilization Project (HCUP)\footnote{\url{https://hcup-us.ahrq.gov/toolssoftware/ccs/AppendixASingleDX.txt}}.

\begin{table}[t]
\caption{The 25 pre-defined phenotypes in the MIMIC-III dataset.}
\label{tab:mimic_phenotypes}
\centering
\resizebox{0.99\linewidth}{!}{
  \begin{tabular}{ll}
  \toprule
  \bfseries Phenotype & \bfseries Type \\
  \midrule
Acute and unspecifed renal failure & acute \\
Acute cerebrovascular disease & acute \\
Acute myocardial infarction & acute \\
Cardiac dysrhythmias & mixed \\
Chronic kidney disease & chronic \\
Chronic obstructive pulmonary disease & chronic \\
Complications of surgical/medical care & acute \\
Conduction disorders & mixed \\
Congestive heart failure; nonhypertensive & mixed \\
Coronary atherosclerosis and related & chronic \\
Diabetes mellitus with complications & mixed \\
Diabetes mellitus without complication & chronic \\
Disorders of lipid metabolism & chronic \\
Essential hypertension & chronic \\
Fluid and electrolyte disorders & acute \\
Gastrointestinal hemorrhage & acute \\
Hypertension with complications & chronic \\
Other liver diseases & mixed \\
Other lower respiratory disease & acute \\
Other upper respiratory disease & acute \\
Pleurisy; pneumothorax; pulmonary collapse & acute \\
Pneumonia & acute \\
Respiratory failure; insufficiency; arrest & acute \\
Septicemia (except in labor) & acute \\
Shock & acute \\
  \bottomrule
  \end{tabular}
}
\end{table}

\section{Baselines}
\label{apd:baselines}
We consider the following baselines in this work:
\begin{itemize}
    \item Logistic Regression~(LR,~\citet{keyhani2008electronic}): It first transforms each EHR visit into a multi-hot vector, and then uses a linear layer to perform prediction.
    \item XGBoost~\citep{chen2016xgboost}: It optimizes the model's performance through gradient descent and regularization techniques. 
    \item Transformer~\citep{li2020behrt}: It directly uses a self-attention structure for modeling EHR visits.
    \item  PT-FT~\citep{pmlr-v193-xu22a}: It adopts the standard pretrain-then-finetune pipeline, which first pretrain on patients with basic features, then finetune on patients with extra features without any other training strategies.
    \item Reweight \citep{li2023multi}: It adaptively adjusts the weight between the patients between basic and extra features.
    \item AUX-TS \citep{han2021adaptive}: It uses the cosine similarity between gradients of loss to balance the weight between basic and extra features.
    \item G-Adv \citep{dai2022graph,liu2021subtype}: It uses  adversarial training during the fine-tuning stage to improve the model's robustness. 
    \item ForkMerge \citep{jiang2023forkmerge}: It is the most recent work on transfer learning, which forks the model into multiple branches and dynamically merges branches to enhance auxiliary-target generalization. To adapt it to our setting, we set two branches to encode the update in pretraining and fine-tuning, respectively.
\end{itemize}

{We recognize that traditional missing data imputation (MDI) approaches seem to be applicable as baselines. However, we highlight the significant gap between MDI problems and our setting, which involves learning with both basic and extra features: (1) MDI approaches often make strong assumptions about the data distribution (e.g., missing at random), whereas in our setting, for patients with only basic features, the extra features are entirely missing.
(2) MDI approaches typically lack the flexibility to handle mixed data types, including both continuous and categorical variables, as found in the EHR data used in our study.
(3) In our setting, the volume of missing values is too substantial for missing data imputation methods to be feasible. Taking the UK Biobank as an example, according to Table~\ref{tab:dataset_stats}, 1,140 out of 1,629 patients have missing values for the entire 1,371 out of 2,013 features. Given such a large proportion of missing data, applying MDI techniques would introduce significant bias in the imputed values.}

{Consequently, we believe that these traditional approaches are not directly adaptable to our setting without significant modifications.}

\section{{Long-term Impact}}
{{\ours} introduces a novel approach to EHR modeling with several potential long-term impacts. Firstly, it enables more comprehensive and personalized healthcare decision-making by seamlessly integrating diverse patient data, including basic and additional features from local medical institutions. This could drive advancements in real-world clinical decision-making, healthcare delivery, and improved patient outcomes through better utilization of rich patient data. Additionally, it inspires future research into transfer learning and domain adaptation techniques for effective knowledge transfer across different patient populations. Moreover, it may also inspire follow-up work such as incorporating different data modalities, enhancing scalability, or applying the approach to other datasets.}

\end{document}